\documentclass[a4paper, 10pt, conference]{ieeeconf}      
\usepackage{FG2023}
\usepackage{multirow}
\usepackage{tabularx,ragged2e,booktabs,babel}
\usepackage{color, colortbl}
\usepackage{graphicx}
\definecolor{Inter}{rgb}{0.819, 0.925, 0.815}
\definecolor{Intra}{rgb}{1.00, 0.894, 0.631}
\definecolor{Multi}{rgb}{0.62, 0.823, 0.843}
\usepackage{amsmath} 
\usepackage{dblfloatfix}
\usepackage{cite}

\FGfinalcopy 

\IEEEoverridecommandlockouts                              
\def\FGPaperID{6} 

\title{\LARGE \bf
Towards Intercultural Affect Recognition:\protect\\Audio-Visual Affect Recognition in the Wild Across Six Cultures
}

\author{\parbox{16cm}{\centering
    {\large Leena Mathur$^1$, Ralph Adolphs$^2$, and Maja J Matarić$^1$ }\\
    {\normalsize
    $^1$ University of Southern California,
    $^2$ California Institute of Technology}}
}

\begin{document}
\ifFGfinal
\thispagestyle{empty}
\pagestyle{empty}
\else
\author{Anonymous FG2023 submission\\ Paper ID \FGPaperID \\}
\pagestyle{plain}
\fi
\maketitle

\begin{abstract}
In our multicultural world, affect-aware AI systems that support humans need the ability to perceive affect across variations in emotion expression patterns across cultures. These systems must perform well in cultural contexts without annotated affect datasets available for training models. A standard assumption in affective computing is that affect recognition models trained and used within the same culture (\textit{intracultural}) will perform better than models trained on one culture and used on different cultures (\textit{intercultural}). We test this assumption and present the first systematic study of intercultural affect recognition models using videos of real-world dyadic interactions from six cultures. We develop an attention-based feature selection approach under temporal causal discovery to identify behavioral cues that can be leveraged in intercultural affect recognition models. Across all six cultures, our findings demonstrate that intercultural affect recognition models were as effective or more effective than intracultural models. We identify and contribute useful behavioral features for intercultural affect recognition; facial features from the visual modality were more useful than the audio modality in this study's context. Our paper presents a proof-of-concept and motivation for the future development of intercultural affect recognition systems, especially those deployed in low-resource situations without annotated data. 

\end{abstract}

\section{INTRODUCTION}
Advances in affective computing, multimodal machine learning, and social signal processing are enabling the development of automated systems that can sense, perceive, and respond to human affective states \cite{baltruvsaitis2018multimodal, picard2000affective, vinciarelli2009social}. \textit{Affect} refers to neurophysiological states that can function as components of emotions or longer-term moods. Psychologists and neuroscientists view affective states as  latent variables that must be inferred from measured variables across communication modalities, such as facial expressions, body postures, and tone of voice  \cite{adolphs2018investigating, barrett2019emotional}. 
Affect is typically represented along two dimensions: how pleasant or unpleasant each state is (valence) and how passive or active each state is (arousal) \cite{russell1980circumplex}. AI systems that are \textit{affect-aware}, possessing the ability to estimate human affect, can enhance the ability of robots and virtual agents to  support human health, education, and well-being \cite{calvo2015oxford, gordon2016affective, rudovic2018personalized}.  

While affect-aware AI systems have the potential to help humans, current affect recognition approaches do not perform well across different cultures; inaccuracies are an issue, particularly, for cultures with limited data available for training affect recognition models. This key, underexplored challenge in affective computing is thought to arise due to differences across cultures in behavioral cues and expression norms that indicate affective states (e.g., facial movements, voice tone) \cite{camras2006culture, jack2012facial, LIM2016105}. Collecting large amounts of  video data across cultures and annotating this data for affect can be expensive, time-consuming, and, for underrepresented cultures, infeasible. This motivates the need for affect recognition approaches that can adapt and perform well across cultures without annotated affect data available.

We address this challenge by developing audio-visual affect recognition models with \textit{attention-based feature selection} (ABFS) under temporal causal discovery \cite{nauta2019causal}. Our approach can be viewed as feature-based unsupervised domain adaptation \cite{kouw2019review} that identifies potential causal feature relationships to be used by models for affect recognition in cultures on which they are not trained. Our approach is motivated by the idea that potential causal relationships between affect and behavioral cues (facial and vocal) in one cultural context may be robust to spurious culture-specific noise, allowing an affect recognition model with ABFS to perform well in cultures on which it is not trained. 

We conducted experiments with SEWA, the largest publicly-available multicultural affect video dataset \cite{kossaifi2019sewa}, that contains real-world dyadic interactions of participants from 6 cultures: British, Chinese, German, Greek, Hungarian, and Serbian. This paper presents the first systematic study of \textit{intercultural} affect recognition models that are trained on videos from one culture and tested on videos from different cultures. We compare the performance of these \textit{intercultural} models to \textit{intracultural} affect recognition models that are trained and tested on videos from the same culture. 

Given the existence of culture-specific emotion expression patterns \cite{camras2006culture, jack2012facial, LIM2016105}, it might be expected that \textit{intracultural} affect recognition models will perform better than \textit{intercultural} models. We test this assumption in our research. Our findings across all six cultures, surprisingly, suggest that intercultural affect recognition models may be as effective and, in some domains, more effective than intracultural models. Our results contribute new baseline findings and a proof-of-concept for the potential of creating \textit{intercultural affect recognition} systems that can be used across cultures. This work makes the following contributions: 
\begin{itemize}
\item The first systematic study of audio-visual intercultural affect recognition models, contributing a new baseline. 
\item New findings regarding the potential of intercultural affect recognition models to match or outperform intracultural models.
\item Analysis and identification of automatically-selected interpretable features, particularly facial cues, that were useful for affect recognition across cultural domains.  
\end{itemize}

\section{RELATED WORKS}
\subsection{Culture and Affect Expression}
Humans within a \textit{culture} typically share common beliefs, values, and social norms that can influence their perception of the world and communication patterns \cite{betancourt1993study}. Scientists since the 1800s \cite{darwin1998expression} have debated the extent to which processes for affect expression are \textit{culturally-universal} versus \textit{culture-specific}. A leading historical perspective is that affect expression patterns tend to be culturally-universal. Studies from participants in Argentina, Brazil, Chile, Japan, New Guinea, and the United States \cite{ekman1971universals, ekman1973cross, ekman1993facial, ekman1971facial, ekman1969pan} identified Facial Action Unit (FAU) configurations that were commonly observed across cultures when participants conveyed basic discrete emotions (e.g., happy, sad). However, these studies did not consider the existence of culture-specific display rules that influence whether it is appropriate, in a given cultural context, to display affective information. Subsequent studies have not supported the view that affect expression patterns are culturally-universal \cite{gendron2014perceptions}. Psychology studies have found that affect expression patterns in face and eye movements vary within cultures and vary even more across cultures \cite{camras2006culture, jack2012facial, scherer2007facial}. Automated analysis of affect expression from images in the wild found very few facial affect expression patterns shared across cultures \cite{srinivasan2018cross}. Since affect expression is influenced by cognitive appraisal mechanisms \cite{arnold1960emotion, lazarus1968emotions}, researchers have posited that differences in cultural value systems and norms may result in culture-specific cognitive appraisal mechanisms for expressing affect \cite{scherer2009culture}. 

\subsection{Continuous Affect Recognition Models Across Cultures}
Given the culture-specific aspects of affect expression patterns, researchers in affective computing have started to acknowledge the importance of creating affect recognition models that can perform well across cultures; this area remains an open research problem \cite{7865884}. A seminal data collection effort to address this problem occurred through the creation of the SEWA database, the first and largest publicly-available video dataset of affect in the wild, including six cultures \cite{kossaifi2019sewa}. We use SEWA data in this paper.

Prior work using a subset of the SEWA database trained audio-visual affect recognition Long Short-Term Memory (LSTM) networks on videos of German participants to predict valence and arousal in videos of Hungarian participants \cite{ringeval2018avec}. Subsequent research efforts leveraged semi-supervised learning on a larger subset of the SEWA database to train recurrent models on videos from German and Hungarian participants and test these models on Chinese participants \cite{mallol2020investigation}. Another research effort \cite{ringeval2019avec} trained shallow LSTM networks on videos from German and Hungarian participants and tested on Chinese participants; this study found that facial features, specifically FAUs, were useful for predicting valence and arousal. These findings motivated us to analyze the contributions of FAU in our paper's experiments. Prior researchers have also investigated elastic weight consolidation for intercultural affect recognition across French and German cultures \cite{han2021internet}, with German data from a subset of the SEWA database and French data from the RECOLA database \cite{ringeval2013introducing}. These prior works did not use data from all six cultures and did not explore methods for selecting useful features for affect recognition across cultures. Our paper, to the best of our knowledge, contributes the first  study of intercultural affect recognition models across all six cultures in SEWA, as well as the first study using ABFS approaches to identify effective behavioral cues for intercultural, audio-visual affect recognition.

\subsection{Causal Discovery in Time-Series Data}
Continuous affect recognition is typically viewed as a multivariate time-series prediction problem, with the goal of predicting affect labels at each timestep of a video. Models that rely on causal feature representations have the potential to capture the underlying dynamics of a domain and to be robust to spurious noise, making them useful for tasks that involve transferring knowledge across domains \cite{scholkopf2021toward}  (e.g., intercultural affect recognition). Various approaches exist for inferring causal features from time-series samples, including constraint-based methods that estimate a single causal graph through conditional independence testing \cite{spirtes2000causation}, score-based methods that optimize a chosen objective when learning a causal graph \cite{bengio2019meta}, and causal graph estimation methods that use gradient-based learning to estimate a causal graph \cite{lachapelle2019gradient, wu2020discovering, nauta2019causal}. We adapt an attention-based causal graph estimation method \cite{nauta2019causal} to identify features that can be useful for affect recognition across cultures. Our approach is motivated by the idea that
potential causal relationships between affect and
behavioral cues in one culture may
be robust to spurious culture-specific noise, supporting the feasibility of intercultural affect recognition models. To the best of our knowledge, we contribute the first study of causal graph estimation approaches for identifying useful features in continuous intercultural affect recognition. 

\begin{figure*}[h]
    \centering
    	\includegraphics[width=.7\paperwidth]{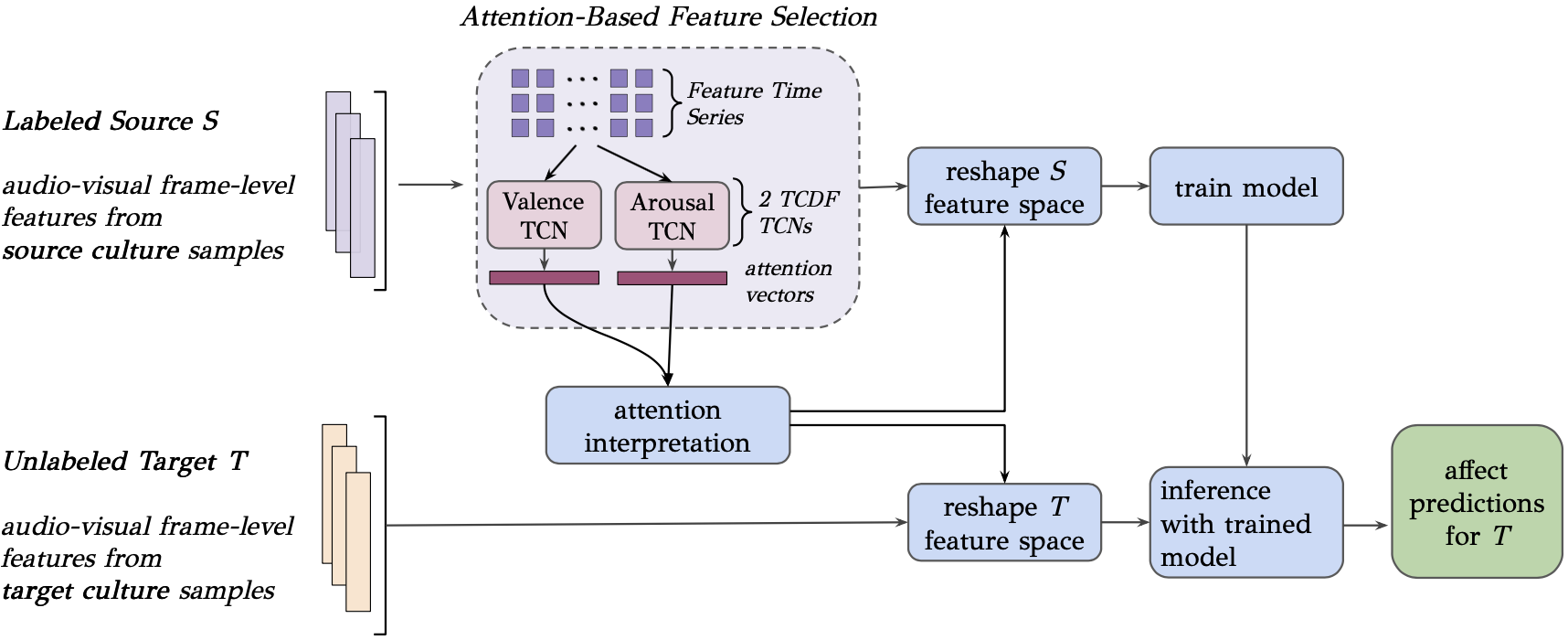}
    	\caption{Visualization of affect recognition models across culture domains with Attention-Based Feature Selection (ABFS).}
    	\label{fig:method}
\end{figure*}

\section{METHODOLOGY}
We conducted a systematic study of intercultural video-based affect recognition with ABFS; an overview of the modeling approach is visualized in \textbf{Fig. \ref{fig:method}}.

\subsection{Multicultural Affect Video Dataset}\label{data}
We used the SEWA Database \cite{kossaifi2019sewa}, the largest video dataset for in-the-wild affect estimation. SEWA contains videos of individuals engaged in naturalistic, real-world dyadic interactions. Aligned with prior studies \cite{kossaifi2020factorized, kossaifi2019sewa, tellamekala2022modelling}, we chose to use the Basic SEWA database (538 video clips, 275 individuals), which contains the most comprehensively annotated subset of videos within SEWA. The individuals ranging from 18-40 years, have a balanced gender representation (52\% male, 48\% female), and come from 6 cultures: British, Chinese, German, Greek, Hungarian, and Serbian. The number of participants varied across cultures within the range of 35 to 56. The valence and arousal of each individual has been labeled at each frame in the videos; the provided annotations are normalized in the continuous range [0, 1]. The final affect labels are an aligned aggregation (canonical time warping \cite{trigeorgis2016deep}) of the frame-level annotations from 5 annotators who shared the same culture as the person they were labeling, boosting the validity of the affect labels. It is worth noting that the ground truth on which models are trained is an aggregation of the annotators’ perceptions of the affect of people in videos; this method for obtaining ground truth is the current norm in affective computing \cite{d2018affective}. Within the Basic SEWA data, we computed the correlation (threshold = 0) of the affect annotations across annotators to identify a final subset of 510 videos. 
\raggedbottom

\subsection{Multimodal Feature Extraction}
We extracted 792 audio-visual features from the speakers in each video frame to represent each speaker’s observable behavioral cues over time. The OpenSMILE toolkit (version 2.0) \cite{eyben2013recent} was used to extract 83 audio features that capture cepstral, spectral, prosodic, energy, and voice quality information from raw speech signals at each audio frame (10 ms step size). 18 audio features came from the GeMAPS feature set, and 65 features came from the ComParE feature set; prior research found these signals to be useful at capturing affective properties of speech \cite{eyben2015geneva, schuller2013interspeech, kossaifi2019sewa}, motivating their use in our work. The OpenFace toolkit14 (version 2.2.0) \cite{baltrusaitis2018openface} was used to extract 709 visual features that capture eye gaze, FAUs, and head pose information from speakers from each video frame. After obtaining these audio-visual features, we aligned audio features, visual features, and affect labels at each video frame. Similar to prior research \cite{tavabi2019multimodal}, we focused on using interpretable audio-visual features from OpenSMILE and OpenFace, instead of using deep feature representations, to more effectively identify and analyze behavioral cues in our experiments.  

\begin{table*}[b]
\centering
 \caption{Results (RMSE/CCC) from Intracultural, Intercultural, and Multicultural Affect Recognition Models.\\ (Intracultural results are in Yellow on the diagonal; Green highlights indicate that the intercultural model was equivalent to or outperformed the corresponding intracultural model; Multicultural results are in Blue).}
\begin{tabular}{clccccccl}
\cline{2-9}
\multicolumn{1}{l|}{} &  & \multicolumn{6}{c}{\textbf{Target Culture}} & \multicolumn{1}{l|}{} \\ \hline
\multicolumn{1}{|l|}{} & \multicolumn{1}{l|}{} & \multicolumn{1}{l|}{\textbf{British}} & \multicolumn{1}{l|}{\textbf{Chinese}} & \multicolumn{1}{l|}{\textbf{German}} & \multicolumn{1}{l|}{\textbf{Greek}} & \multicolumn{1}{l|}{\textbf{Hungarian}} & \multicolumn{1}{l|}{\textbf{Serbian}} & \multicolumn{1}{l|}{\textbf{All}} \\ \cline{2-9} 
\multicolumn{1}{|c|}{\textbf{}} & \multicolumn{1}{l|}{\textbf{British}} & \multicolumn{1}{c|}{\cellcolor{Intra}0.39 / 0.06} & \multicolumn{1}{c|}{\cellcolor{Inter}0.23 / -0.06} & \multicolumn{1}{c|}{\cellcolor{Inter}0.23 / 0.05} & \multicolumn{1}{c|}{\cellcolor{Inter}0.27 / 0.04} & \multicolumn{1}{c|}{\cellcolor{Inter}0.38 / -0.03} & \multicolumn{1}{c|}{0.21 / -0.01} & \multicolumn{1}{l|}{} \\ \cline{2-8}
\multicolumn{1}{|c|}{\textbf{}} & \multicolumn{1}{l|}{\textbf{Chinese}} & \multicolumn{1}{c|}{\cellcolor{Inter}0.36 / -0.03} & \multicolumn{1}{c|}{\cellcolor{Intra}0.25 / 0.02} & \multicolumn{1}{c|}{\cellcolor{Inter}0.24 / -0.03} & \multicolumn{1}{c|}{\cellcolor{Inter}0.27 / -0.02} & \multicolumn{1}{c|}{\cellcolor{Inter}0.37 / 0.01} & \multicolumn{1}{c|}{0.21 / 0.00} & \multicolumn{1}{l|}{} \\ \cline{2-8}
\multicolumn{1}{|c|}{\textbf{}} & \multicolumn{1}{l|}{\textbf{German}} & \multicolumn{1}{c|}{\cellcolor{Inter}0.35 / 0.01} & \multicolumn{1}{c|}{\cellcolor{Inter}0.22 / -0.01} & \multicolumn{1}{c|}{\cellcolor{Intra}0.25 / 0.01} & \multicolumn{1}{c|}{\cellcolor{Inter}0.27 / 0.01} & \multicolumn{1}{c|}{\cellcolor{Inter}0.37 / 0.00} & \multicolumn{1}{c|}{\cellcolor{Inter}0.20 / 0.00} & \multicolumn{1}{l|}{} \\ \cline{2-8}
\multicolumn{1}{|c|}{\textbf{Source Culture}} & \multicolumn{1}{l|}{\textbf{Greek}} & \multicolumn{1}{c|}{\cellcolor{Inter}0.35 / 0.02} & \multicolumn{1}{c|}{\cellcolor{Inter}0.21 / -0.03} & \multicolumn{1}{c|}{\cellcolor{Inter}0.23 / 0.02} & \multicolumn{1}{c|}{\cellcolor{Intra}0.31 / 0.02} & \multicolumn{1}{c|}{\cellcolor{Inter}0.37 / -0.01} & \multicolumn{1}{c|}{\cellcolor{Inter}0.20 / 0.00} & \multicolumn{1}{l|}{} \\ \cline{2-8}
\multicolumn{1}{|c|}{\textbf{}} & \multicolumn{1}{l|}{\textbf{Hungarian}} & \multicolumn{1}{c|}{\cellcolor{Inter}0.36 / -0.03} & \multicolumn{1}{c|}{\cellcolor{Inter}0.25 / 0.02} & \multicolumn{1}{c|}{\cellcolor{Inter}0.24 / -0.02} & \multicolumn{1}{c|}{\cellcolor{Inter}0.30 / -0.03} & \multicolumn{1}{c|}{\cellcolor{Intra}0.51 / 0.05} & \multicolumn{1}{c|}{0.21 / 0.00} & \multicolumn{1}{l|}{} \\ \cline{2-8}
\multicolumn{1}{|c|}{\textbf{}} & \multicolumn{1}{l|}{\textbf{Serbian}} & \multicolumn{1}{c|}{\cellcolor{Inter}0.35 / 0.00} & \multicolumn{1}{c|}{\cellcolor{Inter}0.22 / 0.00} & \multicolumn{1}{c|}{\cellcolor{Inter}0.22 / 0.00} & \multicolumn{1}{c|}{\cellcolor{Inter}0.27 / 0.00} & \multicolumn{1}{c|}{\cellcolor{Inter}0.37 / 0.00} & \multicolumn{1}{c|}{\cellcolor{Intra}0.20 / 0.00} & \multicolumn{1}{l|}{} \\ \cline{2-9} 
\multicolumn{1}{|c|}{\textbf{}} & \multicolumn{1}{l|}{\textbf{All}} & \multicolumn{1}{l}{} & \multicolumn{1}{l}{} & \multicolumn{1}{l}{} & \multicolumn{1}{l}{} & \multicolumn{1}{l}{} & \multicolumn{1}{l|}{} & \multicolumn{1}{c|}{\cellcolor{Multi}0.30 / 0.00} \\ \hline
\multicolumn{1}{l}{} &  & \multicolumn{1}{l}{} & \multicolumn{1}{l}{} & \multicolumn{1}{l}{} & \multicolumn{1}{l}{} & \multicolumn{1}{l}{} & \multicolumn{1}{l}{} & 
\end{tabular}
\label{tab:results}
\end{table*}

\subsection{Problem Formulation for Intercultural Affect Recognition}
\label{methodologydesc}
We approach intercultural affect recognition as a feature-based unsupervised domain adaptation (UDA) problem \cite{kouw2019review, wilson2020survey}. This type of UDA involves training models on labeled source domains and re-shaping the feature space in order to perform tasks on unlabeled target domains. In this paper, a domain refers to a set of videos from a single culture. 

Given a set of source sample videos $S$ with affect labels at each frame and a set of unlabeled target sample videos $T$, the goal is to train a model on $S$ that performs well at predicting affect at each frame of $T$ with minimal error when $S$ and $T$ come from different cultures. Each sample in $S$ and $T$ is padded to the same length of $L=1000$ and has the same initial number of features $D=792$. Each \textit{S} contains \textit{N} time-series representations of $D$-dimensional videos, designated as $SV_1...SV_N$, where each $SV_{1...N} \in R^{L X D}$ and each $SV$ has corresponding affect time-series labels for valence and arousal, continuous between [0, 1]. Each \textit{T} contains \textit{M} unlabeled video time-series representations designated as $TV_1...TV_M$, where each $TV_{1...M} \in R^{L X D}$. We use ABFS (described in \textbf{Section \ref{abfs}}) on samples in $S$ to identify potential causal relationships between the features in $S$ and labels in $S$. We then reshape the feature space of both $S$ and $T$ to include these features. After training our final model on the re-shaped samples of $S$, we perform inference with the trained model on the re-shaped samples of $T$ to obtain affect predictions for the unlabeled $T$ samples. 
\subsection{Attention-Based Feature Selection}
\label{abfs}
We develop an ABFS approach under the temporal causal discovery framework (TCDF) \cite{nauta2019causal} to identify potential causal relationships in each source culture between the audio-visual time-series features and valence and arousal affect labels. Across source samples, we train two separate temporal convolutional neural networks (TCNs), the Valence TCN and the Arousal TCN, to predict valence and arousal at each video frame by using only the past values of the audio-visual features until that frame. Similar to the original paper \cite{nauta2019causal} we fix TCN hyperparameters to make network initialization  constant across all experiments, reported for reproducibility (epochs=1000, kernel size=250, mean square error loss, adam optimizer, dilation coefficient=250).

Each TCN includes a separate channel for each audio-visual time series and includes an attention mechanism that computes attention scores for each audio-visual feature. The Valence TCN and Arousal TCN have trainable attention vectors that are element-wise multiplied by each audio-visual time series feature. Therefore, the final attention vectors for each TCN capture how much attention that TCN pays to each audio-visual feature when predicting valence or arousal. Similar to \cite{nauta2019causal}, we apply a discrete threshold (of 0.25) on these final attention scores to identify potentially causal features. We refer to these selected ABFS features as \textit{potentially causal} so that our research does not make ungrounded assumptions about causality in the studied affect context. Further details regarding the TCDF framework are found in  \cite{nauta2019causal}. We combine  ABFS features selected from the Valence TCN and Arousal TCN into the final feature set used to reshape the source and target domains in intercultural affect recognition, as described in \textbf{Section \ref{methodologydesc}}.

\subsection{Model Training and Metrics}
Using the 6 cultures present in the SEWA database, we conduct 30 experiments with \textit{intercultural} affect recognition models that are trained on videos from one culture and tested on videos from a different culture. Our \textit{intercultural} affect recognition models are baselined against 6 \textit{intracultural} models that are trained and tested on videos from the same culture. For a point of comparison, similar to \cite{kossaifi2019sewa}, we trained and tested a \textit{multicultural} model on groups of videos that contained a random mixture of all cultures. 

All 37 experiments were conducted with 5-fold cross-validation. Within each cross-validation experiment, we standardized the features in both the training and testing set by the distributions in the training set. Aligned with prior SEWA research \cite{ringeval2019avec}, all models were LSTM networks \cite{schmidhuber1997long}, chosen for their potential to capture nonlinear dependencies in time-series data. Similar to prior SEWA research \cite{ringeval2019avec}, we fix LSTM hyperparameters to make network initialization constant across all 37 experiments, reported for reproduciblity (dropout = 0.1, learning rate = 0.1, hidden units = 256, adam optimizer, epochs = 1000 with early stopping). Models were implemented in PyTorch \cite{paszke2017automatic}. 

For each experiment, two metrics were computed at each cross-validation fold to evaluate the model's affect predictions with respect to ground truth: (1) RMSE, the Root mean square error minimized between predictions and ground truth; (2) CCC, the concordance correlation coefficient. Similar to prior SEWA research \cite{toisoul2021estimation}, we trained models with RMSE and CCC loss functions. We used RMSE to compare the prediction performances of the models. 

\section{RESULTS AND DISCUSSION}
\begin{figure*}[h]
    \centering
    	\includegraphics[width=.8\paperwidth]{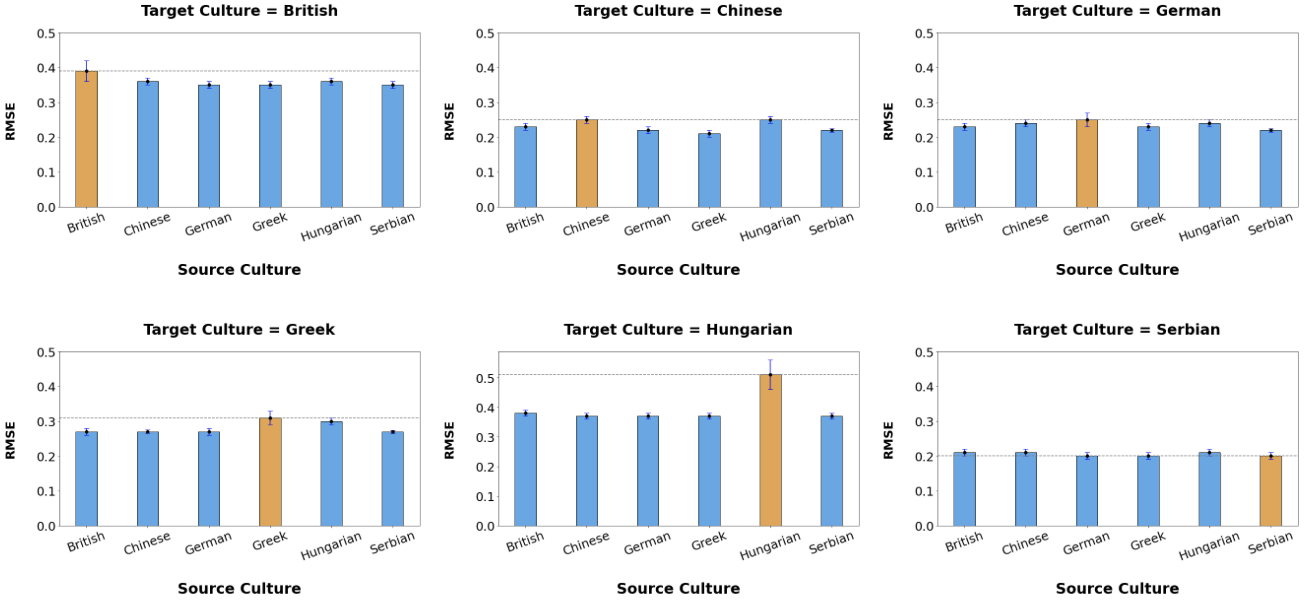}
    	\caption{Comparison of intercultural affect recognition performance (RMSE) with intracultural affect recognition performance (intracultural results are in orange, intercultural results are in blue, error bars are displayed)}
    	\label{fig:model_results}
\end{figure*}

Results from the affect recognition experiments are presented in \textbf{Table \ref{tab:results}}. Our analysis focuses on the RMSE metric. A comparison of intercultural affect modeling results (RMSE) relative to intracultural results for each target culture are visualized in \textbf{Fig. \ref{fig:model_results}}. For 5 of the 6 target cultures, British, Chinese, German, Greek, and Hungarian, each intercultural affect recognition model outperformed or matched the corresponding intracultural model. In the Serbian case, the intercultural models closely matched the performance of the intracultural model. We found that 27/30 intercultural models were equivalent to or outperformed their intracultural models, and 19/30 intercultural models outperformed the multicultural model. \textit{Our findings suggest that intercultural affect modeling may be as effective and, in some domains, more effective than intracultural modeling.}  

The intercultural affect recognition approach with ABFS appears to have selected features within each culture that were less-influenced by culture-specific noise (e.g., display rules \cite{matsumoto1990cultural}). To identify the key behavioral cues that enabled intercultural affect recognition models to transfer knowledge across cultures and outperform intracultural models, we examined the ABFS features selected by each model. \textbf{Table \ref{tab:my_label}} lists these features, along with feature descriptions. Across all source cultures, all of the selected ABFS features were from different feature sets within the visual modality. It appears the visual modality was more useful than the audio modality for affect recognition in the studied SEWA context. Within the 15 selected features from the visual modality, there were 4 features from PDM parameters (face shape representations), 2 features from head pose movements,  4 features from eye and eyebrow movements, and 5 features from lip, nose, and cheek movements. These findings on feature importance align with a prior study \cite{ringeval2019avec} that found FAU features important for affect recognition with a smaller subset of the SEWA dataset. Our findings indicate that temporal patterns in facial movements have potential for use in intercultural affect recognition models.

\begin{table}[h]
\centering
    \caption{Features Selected by the ABFS  Models.}
    \begin{tabular}{l|l}
        \textbf{Source }&\textbf{ABFS Features Selected and Used}\\
        \hline
         British & PDM parameter 10 (deformation due to expression)\\
        \hline
         Chinese & Head pose roll in the Z axis (pose Rz)\\
         & Left eye gaze direction, x coordinate (Gaze 0 X)\\
        \hline
        & Intensity lip tightener (FAU 23)\\
        & Left lip landmark (X 60)\\
         German & Left cheek landmark (X 1)\\
        \hline
         Greek & PDM parameters 2 (deformation due to expression)\\
         & PDM parameter 27 (deformation due to expression)\\
         & Intensity of inner brow raise (FAU 1)\\
         & Head pose roll along the Z axis (pose Rz) \\
        \hline
         Hungarian & nose landmark (X 30, moving with head movement)\\
         & Lip corner pull intensity (FAU 12)\\
         & Eye gaze direction for right eye (Gaze 1 X)\\
        \hline
         Serbian & PDM parameter 16 (deformation due to expression)\\
         & Right eye landmark (X 44)
         
    \end{tabular}
    \label{tab:my_label}
\end{table}

Our analysis focused on RMSE, a standard metric for time-series prediction, to compare the prediction performance of intercultural and intracultural affect recognition models. We report CCC for each experiment for consistency with prior papers that used subsets of SEWA and included CCC \cite{kossaifi2019sewa, ringeval2019avec, ringeval2018avec}. We contribute new RMSE and CCC findings for intercultural affect recognition across all six cultures in the SEWA data. Prior works focused on only two or three of the six cultures, each leveraging different training, validation, and testing splits. These differences constrained our ability to make direct comparisons with prior papers. Low CCC in our models motivates future work to create affect recognition techniques that perform well in minimizing regression error and maximizing correlation metrics across cultures.

\section{CONCLUSION}
Our paper presents the first systematic study of audio-visual intercultural affect recognition models, using videos of real-world dyadic interactions from six cultures. We contribute new findings regarding the potential for intercultural affect recognition models to match or outperform intracultural models. Our results serve as a baseline, proof-of-concept, and motivation for the future development of intercultural affect recognition systems that can be deployed in cultures without annotated affect datasets for training. Through our ABFS approach, we identified and analyzed the automatically-selected features (primarily facial cues) that were useful for affect recognition across cultures; these findings inform future approaches for video-based intercultural affect recognition. 

To support the robustness and generalizability of intercultural affect recognition findings, future work could include collecting and experimenting with larger multicultural affect datasets to encompass participants, culture domains, and annotators beyond the six cultures in SEWA. Future work might also venture beyond valence and arousal to explore models for jointly recognizing higher-level affective states (e.g., "happy") in the design of intercultural human-machine interaction systems. Our paper informs and motivates the future development of affect-aware AI systems to support humans in the wild across cultures.

%
\section{ACKNOWLEDGMENTS}
This work was supported by a Caltech Summer Undergraduate Research Fellowship. The research in this paper uses the SEWA Database collected in the scope of SEWA project financially supported by the European Community’s Horizon 2020 Programme (H2020/20142020) under Grant agreement No. 645094.

{\small
\bibliographystyle{ieee}
\bibliography{egbib}
}

\end{document}